\documentclass[conference]{IEEEtran}
\IEEEoverridecommandlockouts  

\usepackage{cite}
\usepackage{amsmath,amssymb,amsfonts}
\usepackage{algorithmic}
\usepackage{graphicx}
\usepackage{booktabs}
\usepackage{siunitx}  
\usepackage{amsmath}
\usepackage{algorithm}
\usepackage[table,xcdraw]{xcolor}
\usepackage{threeparttable}
\usepackage{textcomp}
\usepackage{xcolor}
\usepackage{booktabs}
\usepackage{multirow}
\usepackage{hyperref}
\hypersetup{
    colorlinks=true,
    linkcolor=blue,
    urlcolor=blue,
    citecolor=blue
}
\usepackage{pifont} 
\newcommand{\cmark}{\ding{51}} 
\newcommand{\xmark}{\ding{55}} 

\usepackage{comment}
\usepackage{tikz}          
\usetikzlibrary{positioning} 

\title{
    Hyper-V2X: Hypernetworks for Estimating Epistemic and Aleatoric Uncertainty in Cooperative Bird’s-Eye-View Semantic Segmentation
}

\author{Abhishek~Dinkar~Jagtap$^{1\;*}$,
         Sanath Tiptur Sadashivaiah$^{2}$,
         Andreas~Festag$^{1}$
    \thanks{*This  work  was  supported  by  the  German  Science  Foundation  (DFG)  within  the project \emph{Beyond Validation\textsuperscript{AI}} (grant no.  549102058).}
    \thanks{$^{*}$
       Corresponding author: 
       {abhishekdinkar.jagtap@carissma.eu}.%
    }%
    \thanks{$^{1}$
        CARISSMA Institute for Electric, COnnected, and Secure Mobility (C-ECOS), Technische Hochschule Ingolstadt, Ingolstadt, 85049, Germany.
    }%
    \thanks{$^{2}$
        University of Applied Sciences Aschaffenburg, Aschaffenburg, 63743, Germany.
    }%
}

\begin{document}

\pagestyle{empty}

\maketitle

\begin{abstract}

Cooperative perception enabled by Vehicle-to-Everything (V2X) communication enhances autonomous driving safety by creating a unified environmental representation through shared sensory data. While recent works have advanced multi-agent fusion for improved perception, uncertainty quantification in such cooperative frameworks remains largely unexplored. This paper introduces Hyper-V2X, a hypernetwork-based framework for estimating both epistemic and aleatoric uncertainties in V2X-based perception. Specifically, we propose a partial weight generation scheme and V2X context embedding module that conditions a Bayesian hypernetwork on fused multi-agent features to generate weight distributions for stochastic Bird’s-Eye-View (BEV) segmentation. Unlike existing deterministic BEV models, Hyper-V2X enables efficient uncertainty estimation with little computation overhead. Our approach is architecture-agnostic, and can be  seamlessly integrating with modern cooperative backbones such as CoBEVT. Experiments on the OPV2V benchmark demonstrate that Hyper-V2X provides accurate, well-calibrated uncertainty estimates and improves overall perception reliability. Our code and benchmark are publicly available under an open-source license: \url{https://github.com/abhishekjagtap1/Hyper-V2X} 

\end{abstract}
\section{Introduction} 
\label{sec:intro}
Cooperative Perception (CP) has emerged as a key technology for enhancing autonomous driving safety by extending the perceptual range of individual vehicles beyond their own sensor limitations~\cite{Wan_SystematicLitRev_TITS_2025}. This paradigm, facilitated by Vehicle-to-Everything (V2X) communication, enables agents to share information via raw sensor data~\cite{Chen2019_Cooper}, intermediate features~\cite{Hu2022_Where2comm,Yang2023_How2comm}, or object lists through standardized Collective Perception Messages (CPMs)~\cite{Delooz2023_CollPerception, FirstMile}. Consequently, CP has demonstrated significant performance improvements across a spectrum of perception tasks, including cooperative 3D object detection \cite{Hu_CollaborHelps_CVPR_2023, xu2022v2x}, bird's-eye view (BEV) segmentation~\cite{xu2022cobevt, Fu_GenerMaps_CVPR_2025}, 3D occupancy prediction~\cite{song2024_collaborative}, and more recently, collaborative end-to-end driving \cite{liu2024:collaborative} and 3D scene reconstruction \cite{Jagtap_V2XGaussians_2025}. 


Despite these advances, the deployment of deep neural networks (DNNs) for Connected Autonomous Vehicle (CAV) in real-world driving scenarios remains challenging. DNNs are data-hungry and parameter-heavy, requiring large-scale annotated datasets for training -- a particular bottleneck in V2X environments where synchronized, multi-agent data collection is complex and costly. In safety-critical autonomous systems, such data limitations increase the risk of erroneous predictions, emphasizing the need for models that are not only accurate but also capable of expressing how certain they are about their predictions. Therefore, quantifying predictive uncertainty is essential to achieve reliable and trustworthy cooperative perception. To address this challenge, the research community has pursued diverse strategies. Bayesian methods~\cite{krueger2017bayesian}, through techniques like Monte Carlo (MC) Dropout~\cite{pmlr-v48-gal16} or variational inference~\cite{Blundell2015VI}, explicitly model weight distributions. Alternatively, non-Bayesian approaches such as deep ensembles~\cite{NIPS2017_9ef2ed4b} generate multiple predictions to approximate a distribution, while other methods direct a single network to estimate uncertainty parameters directly from the input data.

While the broader field of uncertainty estimation in deep learning is well-established, its application to cooperative perception remains a notably under-explored frontier. \cite{Su2022uncertainty}~proposed a double-quantification method to estimate uncertainty in collaborative object detection. \cite{Huang_UncertCollPerc_Adv_2025} studied the uncertainties in terms of adversarial attacks. To our knowledge, we are the first to apply Hypernetworks in the cooperative perception domain. Specifically, we propose a V2X context embedding module and a partial weight generation strategy that enable efficient uncertainty estimation for cooperative BEV semantic segmentation. 


In this paper, we propose Hyper-V2X, which enables the quantification of uncertainties for Cooperative BEV segmentation and can be employed with little overhead to existing architectures and CP methods. Our main contributions are summarized as follows:

\noindent\textbf{1) Bayesian hypernetwork formulation for cooperative perception:} We introduce a Bayesian hypernetwork that jointly estimates epistemic and aleatoric uncertainties in BEV segmentation with minimal computational overhead.

\noindent\textbf{2) V2X context embedding:} We propose a learnable context embedding that conditions the Hypernetwork on fused multi-agent features enabling context-aware and architecture-agnostic weight generation for predictive uncertainty. 


\noindent\textbf{3) Partial weight generation for Bayesian hypernetworks:} We design a partial weight generation strategy for Bayesian hypernetworks that avoids generating the full set of model parameters, enabling efficient and scalable uncertainty estimation.



\noindent\textbf{4) Comprehensive evaluation on the OPV2V benchmark:} We conduct extensive experiments and ablation studies demonstrating that Hyper-V2X achieves accurate, well-calibrated uncertainty estimates and improves segmentation performance under varying communication conditions.

\begin{figure*}[t!]
   \centering
   \includegraphics[width=\textwidth]{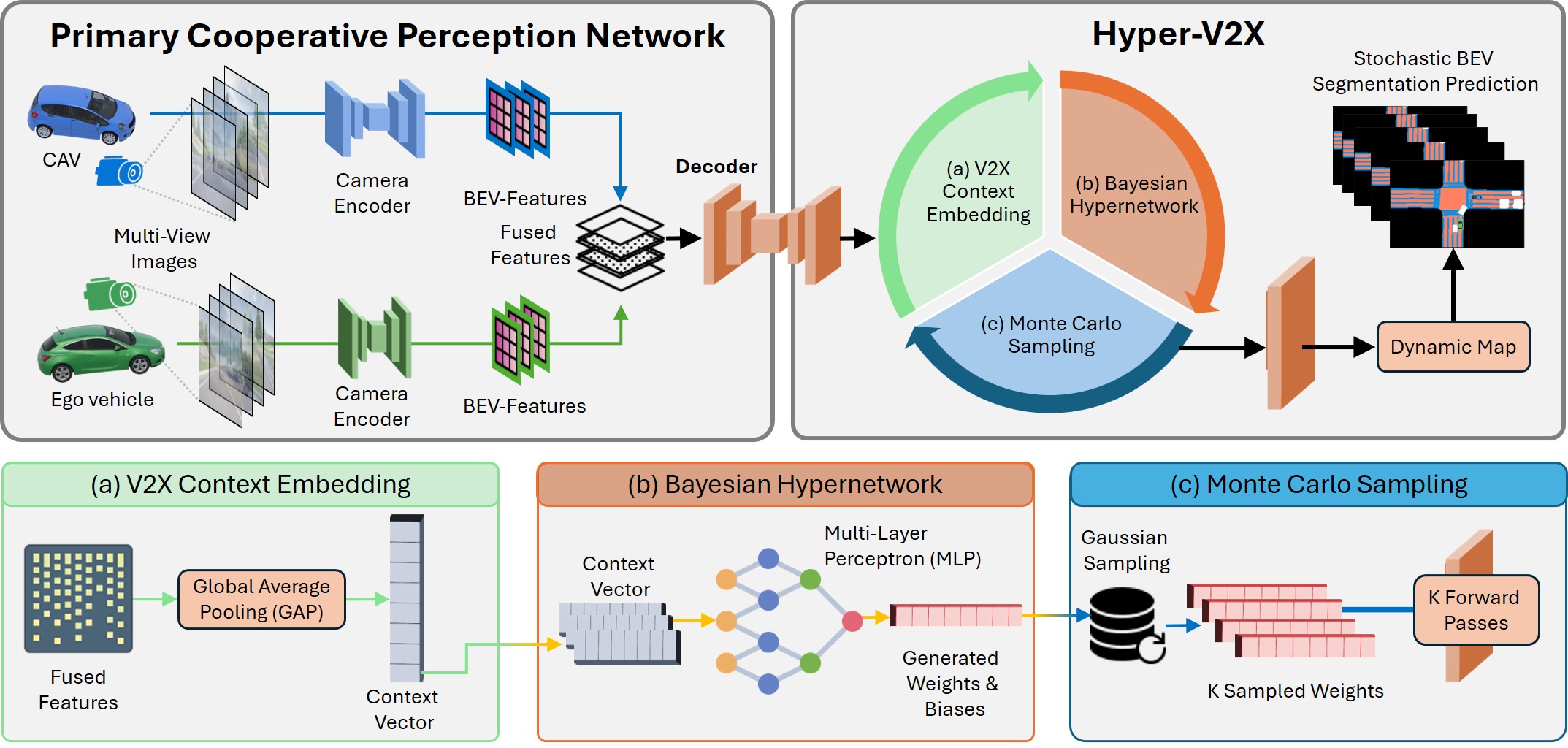}
   \caption{\small Overview of the proposed Hyper-V2X framework for uncertainty estimation in V2X-based cooperative perception.}
   \label{fig:pipeline}
\end{figure*}

\section{Related Work}
\label{sec:related}

\subsection{Cooperative Perception}
Cooperative perception has been widely investigated to address the limited sensing range and occlusions of ego vehicles by sharing perception information among connected vehicles and infrastructure via Vehicle-to-Everything (V2X) communication. Existing approaches are broadly categorized by their fusion strategy into early, intermediate, and late fusion. In early fusion~\cite{Chen2019_Cooper}, raw sensor data such as LiDAR, RADAR point clouds and raw images are transmitted and fused, preserving complete information but incurring prohibitive communication overhead. To mitigate bandwidth constraints, late-fusion~\cite{Delooz2023_CollPerception} methods share only task-level outputs such as detected bounding boxes, reducing communication costs but suffering from information loss due to abstraction. Striking a balance between accuracy and efficiency, intermediate fusion has emerged as a promising direction~\cite{wang2020v2vnet }, where agents exchange compact intermediate feature representations that preserve essential scene context while reducing communication overhead.

\subsection{Uncertainty Estimation}
Trustworthy deep learning necessitates reliable and calibrated uncertainty estimates that align with model prediction accuracy. Traditional methods such as Bayesian Neural Networks (BNNs)~\cite{louizos2017multiplicative}, which use variational inference to learn posterior distributions over model weights, and Deep Ensembles~\cite{NIPS2017_9ef2ed4b}, which train multiple ensemble models explicitly, are widely used for capturing predictive uncertainty. Recently, ~\cite{chan2024estimating} introduces hyper-diffusion models that quantifies different sources of uncertainty by employing a hypernetwork to generate the entire set of model parameters. However, these approaches are computationally expensive, as they require training  multiple ensemble models leading to slow inference and training times. This computational burden becomes a significant challenge in cooperative perception systems, where CP models are typically massive, containing millions of parameters. Moreover, existing approaches for uncertainty quantification in cooperative perception have mainly focused on object detection. \cite{Su2022uncertainty} introduced a double-m-quantification framework for collaborative detection, \cite{Huang_UncertCollPerc_Adv_2025}~studied robustness under adversarial perturbations, and \cite{su2024collaborative} used conformal prediction for object detection and tracking. In contrast, we propose  a hypernetwork-based approach with a partial weight generation strategy to efficiently quantify both epistemic and aleatoric uncertainty for cooperative BEV semantic segmentation.

\subsection{BEV Semantic Segmentation}

BEV semantic segmentation transforms multi-view camera images into a unified top-down representation where each pixel is assigned a semantic category. This representational space has proven transformative for autonomous driving~\cite{li2022bevformer}, providing a natural interface for downstream tasks such as motion planning and trajectory prediction. Subsequent works have explored BEV segmentation for cooperative perception. CoBEVT~\cite{xu2022cobevt} proposed a novel fused axial attention mechanism for efficient feature fusion. BEV-V2X~\cite{Chang2023_BEV-V2X} explores spatio-temporal attention for predicting occupancy grids. CoGMP \cite{Fu_GenerMaps_CVPR_2025} proposes generative priors for efficient compression.
Furthermore, previous studies have introduced uncertainty quantification for BEV segmentation~\cite{Yang2023UncertaintyBEV} enabling autonomous driving systems to evaluate prediction reliability for safer decision-making. However, these methods are limited to single-agent configurations and do not account for the additional complexities introduced by multi-agent collaboration, such as communication noise and feature fusion. In this work, we address this gap by investigating uncertainty quantification in cooperative BEV  semantic segmentation.

\section{Hyper-V2X}
\label{sec:v2xgaussians}

In this section, we introduce the overall framework of Hyper-V2X by expanding on how Bayesian hypernetworks can be integrated into a CP framework for estimating epistemic and aleatoric uncertainty.

\subsection{Preliminaries on Hypernetworks}
\label{sec:hypernetwork}

Hypernetworks~\cite{ha2017hypernetworks} are a class of neural networks that aims to predict the weights/parameters of another neural network, often referred to as the primary or target network. Let the primary network be represented as a function:
\begin{equation}
f_\theta: \mathcal{X} \rightarrow \mathcal{Y}, \quad \hat{y} = f_\theta(x),
\end{equation}

where $x \in \mathcal{X}$ is the input, $\hat{y} \in \mathcal{Y}$ is the output, $\theta \in \Theta$ are the learnable parameters of the primary network, and $\Theta$ denotes the space of all valid parameter configurations. In standard DNN training, the parameters $\theta$ are optimized via backpropagation to minimize a task-specific loss:
\begin{equation}
    \theta^* = \arg \min_{\theta \in \Theta} \mathcal{L}(f_\theta(x), y).
\end{equation}

A Hypernetwork, denoted as $h_\phi$, maps a conditioning vector $c \in \mathcal{C}$ (e.g., Gaussian noise) to the weights of the primary network. As a result the primary network now becomes a function of the Hypernetwork parameters:
\begin{equation}
    \hat{y} = f_{h_\phi(c)}(x).
    \label{eq:hp_params}
\end{equation}

In Eq.~\ref{eq:hp_params}, $h_\phi: \mathcal{C} \rightarrow \Theta$ maps the conditioning space $\mathcal{C}$ to the primary network weight space $\Theta$. During the training, the losses are backpropagated through the primary network to update only the Hypernetwork parameters $\phi$: 
\begin{equation}
    \phi^* = \arg \min_\phi \mathcal{L}\big(f_{h_\phi(c)}(x), y\big).
    \label{eqn:Hypernet}
\end{equation}

Bayesian hypernetworks~\cite{krueger2017bayesian} extend standard Hypernetworks from~Eq.~\ref{eqn:Hypernet} by generating a \emph{distribution} over the primary network weights instead of a single deterministic set. This allows for an ensemble of networks with different parameter configurations, enabling uncertainty estimation in the model predictions. The gradient flow for Bayesian hypernetworks is then given by the expectation of the loss over different sampled weights and is characterized by:
\begin{equation}
    \theta \sim q_\phi(\theta \mid c), \quad c \in \mathcal{C}, \; \theta \in \Theta,
\end{equation}
\begin{equation}
        \mathcal{L}_\text{BHN}(\phi) = 
\mathbb{E}_{\theta \sim q_\phi(\theta \mid c)} \big[ \mathcal{L}(f_\theta(x), y) \big].
\end{equation}

\subsection{Primary Cooperative Perception Network}

Building on the Hypernetwork paradigm for weight generation from Sec.~\ref{sec:hypernetwork}, we propose an alternative approach for training CP models that enables explicit modeling of predictive uncertainty. In our formulation, the CP model is treated as the primary network, whose parameters are dynamically generated by a Hypernetwork. To illustrate this, we present the Hyper-V2X pipeline in the context of a cooperative BEV semantic segmentation task, using CoBEVT~\cite{xu2022cobevt} as our primary CP model as shown in Fig.~\ref{fig:pipeline}. 

Let the overall cooperative BEV segmentation model be represented as $F_\theta: \mathcal{I} \rightarrow \mathcal{Y}$, where $\mathcal{I} = \{I_1, I_2, \dots, I_V\}$ with $I_v \in \mathbb{R}^{H \times W \times 3}$ representing the set of multi-view RGB images from $V$ connected vehicles, and $\mathcal{Y} \in [0, 1]^{H \times W \times C_d}$ denotes the BEV semantic map containing $C_d$ dynamic classes. Subsequently, the learnable parameters $\theta$ of the primary network $F_{\theta}$ can be decomposed into $\theta = \{\theta_\text{enc}, \theta_\text{dec}\}$, where $\theta_\text{enc}$ encompasses the encoder parameters, comprising the shared feature extractor and fusion module following the architecture in \cite{xu2022cobevt}, and $\theta_\text{dec}$ represents the parameters of the decoder head responsible for generating the dynamic BEV semantic map. Using this parameterization, the encoder $F_{\theta_{\text{enc}}}$ produces fused BEV features from all $V$ connected vehicles through:
\begin{equation}
\label{eqn:fused_features}
\mathcal{G} = F_{\theta_{\text{enc}}}(\mathcal{I}, \mathcal{P_{\text{ego}}}) = \left(\bigoplus_{v=1}^{V} \mathcal{T}_{P_{ego}}\left(f_{\theta_{\text{enc}}}(I_V)\right)\right),
\end{equation}
where $f_{\theta_{\text{enc}}}(I_V)$ represents BEV features of each CAV, $\mathcal{T}_{P_v}$ denotes the spatial transformation matrix that maps each CAV's local BEV features to the ego vehicle's coordinate system $P_{\text{ego}}$, 
$\bigoplus$ fuses all transformed features, and $\mathcal{G}$ represents the aggregated BEV features transformed to a unified ego coordinate space. 


\subsection{V2X Context Embedding}
Directly optimizing Hypernetworks to generate the complete parameter set of large-scale models presents significant challenges due to the high-dimensional output space and limited conditioning signal \cite{ha2017hypernetworks, hemati2023partial}. To mitigate this, we adopt a partial weight generation strategy in which the Hypernetwork produces only the decoder parameters $\theta_{\text{dec}}$, while the encoder parameters $\theta_{\text{enc}}$ are directly optimized within the primary network. Conditioning Bayesian hypernetworks solely on random noise often yields weight distributions misaligned with the task-specific manifold, particularly in high-dimensional parameter spaces such as cooperative BEV segmentation.

To address this limitation, we introduce a V2X context embedding module that provides a learnable, task-specific conditioning signal derived from the aggregated BEV features produced by $\theta_{\text{enc}}$.
For each batch instance $b$, the corresponding context embedding $\mathbf{z}_b \in \mathbb{R}^C$ is computed as:
\begin{equation}
   \mathbf{z}_{b,c} = \frac{1}{H \cdot W} \sum_{h=1}^{H} \sum_{w=1}^{W} \mathcal{G}_{b,c,h,w},
\end{equation}
where $\mathcal{G} \in \mathbb{R}^{B \times C \times H \times W}$ denotes the fused BEV feature map obtained from all connected vehicles $I_V$. The resulting global context vector $\mathbf{z}_b$ serves as an adaptive conditioning input for the Bayesian hypernetwork, enabling context-aware and task-specific weight generation for the decoder parameters $\theta_{\text{dec}}$.

\subsection{Bayesian Hypernetwork}
To explicitly model predictive uncertainty in cooperative BEV segmentation, we employ a Bayesian hypernetwork that learns to generate a distribution over the decoder weights of the primary network. Specifically, we replace the decoder with an MLP parameterized by $\phi$. Conditioned on the V2X context embedding for each instance in the data distribution, the Bayesian hypernetwork is trained to predict the mean $(\mu)$ and log-variance $(\log \sigma^2)$ of a Gaussian posterior over the decoder parameters $\theta_{\text{dec}}$. From the resulting Gaussian posterior, we perform MC sampling to draw $K$ sets of decoder weights, where each sample $\theta_{\text{dec}}^{(k)}$ represents a distinct stochastic instantiation of the decoder. Consequently, $K$ forward passes yield $K$ stochastic BEV segmentation predictions:
\begin{equation}
  \theta_{\text{dec}}^{(k)} \sim \mathcal{N}(\mu, \sigma^2),
\end{equation}
\begin{equation}
  \theta_{\text{dec}}^{(k)} = \mu + \sigma \odot \epsilon^{(k)}, \quad \epsilon^{(k)} \sim \mathcal{N}(0, I).
\end{equation}

Together, the Bayesian hypernetwork and MC sampling enable estimation of both epistemic and aleatoric uncertainties, providing a principled framework for reliable uncertainty-aware cooperative perception.
Because the Hypernetwork is conditioned solely on the fused BEV feature representation rather than the backbone architecture, the proposed formulation remains model-agnostic and integrates seamlessly with diverse cooperative perception backbones.


\begin{table}[t]
\centering
\caption{ Bird's-eye-view segmentation results on OPV2V camera-track. We report IoU (\%) for all classes.}
\label{tab:opv2v_segmentation_uncert}
\begin{tabular}{l S[table-format=2.1] S[table-format=2.1] S[table-format=2.1] c}
  \toprule
    \textbf{Method} & \textbf{ Vehicle IoU}  & \textbf{Uncertainty} \\
  \midrule
    No Fusion                        & 37.7 & \xmark \\
    Map Fusion                       & 45.1 & \xmark \\
    F-Cooper~\cite{chen2019fcooper}  & 52.5 & \xmark \\
    AttFuse~\cite{xu2022opv2v}       & 51.9 & \xmark \\
    V2VNet~\cite{wang2020v2vnet}     & 53.5 & \xmark \\
    DiscoNet~\cite{li2021learning}   & 52.9 & \xmark \\
    FuseBEVT~\cite{xu2022cobevt}     & 59.0 & \xmark \\
    CoBEVT~\cite{xu2022cobevt}       & 60.4 & \xmark \\
    \textbf{Hyper-V2X (Ours)}        & \textbf{61.4} & \cmark \\
  \bottomrule
\end{tabular}
\end{table}

\begin{table}[b]
\centering
\caption{Uncertainty estimation under different compression rates. We report IoU (\%) and uncertainty metrics.}
\label{tab:compression_uncertainty}
\begin{tabular}{S[table-format=2.0] S[table-format=3.0] S[table-format=2.2] S[table-format=1.4] S[table-format=1.4] S[table-format=1.4]}
\toprule
{CPR} & {CV (KB)} & {Vehicle IoU $\uparrow$} & {ECE $\downarrow$} & {BS $\downarrow$} & {NLL $\downarrow$}\\
\midrule
0  & 524 & 61.4 & 0.0140 & 0.1618 & 0.5047 \\
2  & 264 & 60.03 & 0.0146 & 0.1764 & 0.5358 \\
4  & 132 & 59.57 & 0.0150 & 0.1789 & 0.5411 \\
8  &  66 & 60.18 & 0.0141 & 0.1820 & 0.5478\\
32 &  16 & 59.50 & 0.0153 & 0.1840 & 0.5519 \\
64 &   8 & 59.48 & 0.0146 & 0.1831 & 0.5501 \\
\bottomrule
\end{tabular}
\end{table}

\subsection{Uncertainty Estimation}

Epistemic uncertainty arises from uncertainty in model parameters~\cite{chan2024estimating}, i.e., how confident the model is in its learned weights. Using MC sampling, a Bayesian hypernetwork generates multiple plausible decoder weights $\theta_{\text{dec}}^{(k)}$, producing $K$ stochastic BEV segmentation predictions from the same fused features $\mathcal{G}$ (Eq.~\ref{eqn:fused_features}). 

The variance across these predictions quantifies epistemic uncertainty:
\begin{equation}
U_{\text{E}} = \frac{1}{C} \sum_{c=1}^{C} \operatorname{Var}_{k} \left[p_c^{(k)}\right],
\label{eq:epistemic}
\end{equation}
where $p_c^{(k)}$ is the softmax probability for class $c$ from the $k$-th sample. 
Aleatoric uncertainty, on the other hand, captures inherent noise in sensor observations and data ambiguity. It is estimated from the entropy of the mean predictive distribution across the $K$ stochastic samples:
\begin{equation}
U_{\text{A}} = - \sum_{c=1}^{C} \bar{p}_c \log \bar{p}_c,
\label{eq:aleatoric}
\end{equation}
where $\bar{p}_c$ represents the mean class probability across all samples.

\section{Experimental Evaluation}
\label{sec:setup-eval}

\begin{figure*}[t]
    \centering
     \includegraphics[width=\linewidth, trim={5mm 5mm 10mm 5mm}, clip]{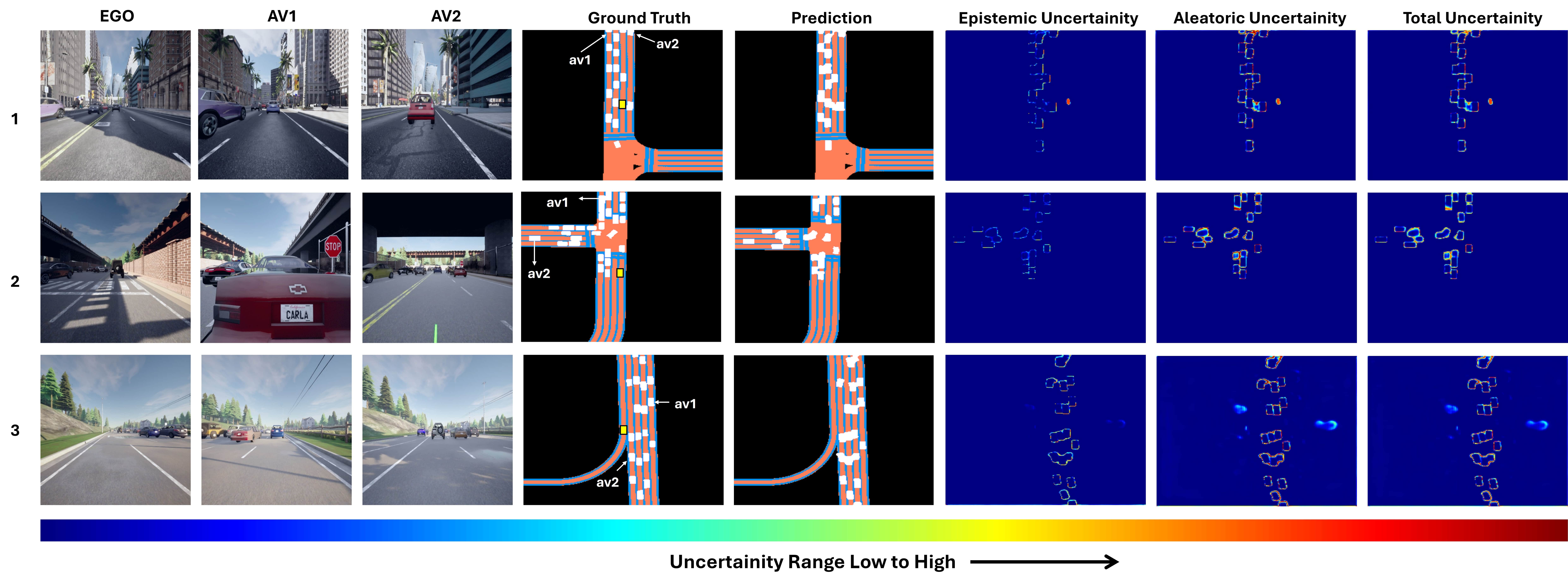}
    \caption{\small \textbf{Qualitative results on OPV2V benchmark.} Ground truth, predicted BEV segmentation, and corresponding epistemic and aleatoric uncertainty maps for representative scenes.}
    \label{fig:visual_res1}
\end{figure*}


This section discusses the dataset, evaluation metrics, and experimental setup. It presents both quantitative and
qualitative results along ablation study on communication data volume.  

\begin{table}[b]
\centering
\caption{Comparison of Uncertainty Estimation Methods.}
\label{tab:uncertainty_methods}
\begin{tabular}{l S[table-format=2.1] S[table-format=1.4] S[table-format=1.4] S[table-format=1.4]}
  \toprule
    \textbf{Method} & \textbf{Vehicle IoU $\uparrow$} & \textbf{ECE $\downarrow$} & \textbf{BS  $\downarrow$} & \textbf{NLL  $\downarrow$}\\
  \midrule
    MC Dropout (0.1)  & 60.3586  & 0.0168  & 0.1424 & 0.4631\\
    MC Dropout (0.3)  & 60.0427 & 0.017  & 0.1450 & 0.4686\\
    MC Dropout (0.5)  & 59.43  & 0.0173  & 0.1460 & 0.4707\\
    MC Dropout (0.7)  & 57.2899  & 0.0179  & 0.1624 & 0.5058\\
    \textbf{Hyper-V2X (Ours)} & \textbf{61.4} & \textbf{0.014} & \textbf{0.1618} & \textbf{0.5047}\\
  \bottomrule
\end{tabular}
\end{table}

\subsection{Dataset and Evaluation Metrics}

We evaluate our approach on the OPV2V dataset~\cite{xu2022opv2v}, a large-scale cooperative perception benchmark. The dataset comprises 73 diverse driving scenarios with an average duration of approximately 25 seconds. Each scenario involves 2–7 connected autonomous vehicles (AVs), each equipped with one LiDAR sensor and four RGB cameras providing a 360° horizontal field of view.  In this work, only the camera data are utilized to perform BEV semantic map prediction around a fixed ego vehicle. The evaluation region covers a $100\,\text{m} \times 100\,\text{m}$ area centered on the ego vehicle, discretized at a $0.39\,\text{m}$ resolution. 
We report Intersection-over-Union (IoU) to assess segmentation performance. 

To evaluate uncertainty, we employ standard calibration and probabilistic metrics. Expected Calibration Error (ECE) measures the alignment between predicted confidence and actual accuracy, Brier Score (BS) quantifies the mean squared error between predicted probabilities and true labels, and Negative Log-Likelihood (NLL) assesses the likelihood of the ground-truth under the predicted distribution~\cite{Pavlitska_2025_ICCV}. The ECE is computed by partitioning predictions into $M$ confidence bins ${B_m}_{m=1}^{M}$ and calculating the weighted average gap between accuracy and confidence:

\begin{equation}
\text{ECE} = \sum_{m=1}^{M} \frac{|B_m|}{N} , \big| \text{acc}(B_m) - \text{conf}(B_m) \big|,
\end{equation}

where $\text{acc}(B_m)$ and $\text{conf}(B_m)$ denote the average accuracy and confidence in bin $B_m$. Together, these metrics provide a comprehensive assessment of both prediction quality and uncertainty calibration.


\label{sec:results_comp}

\begin{figure*}[t]
    \centering
    \includegraphics[width=0.8\linewidth]{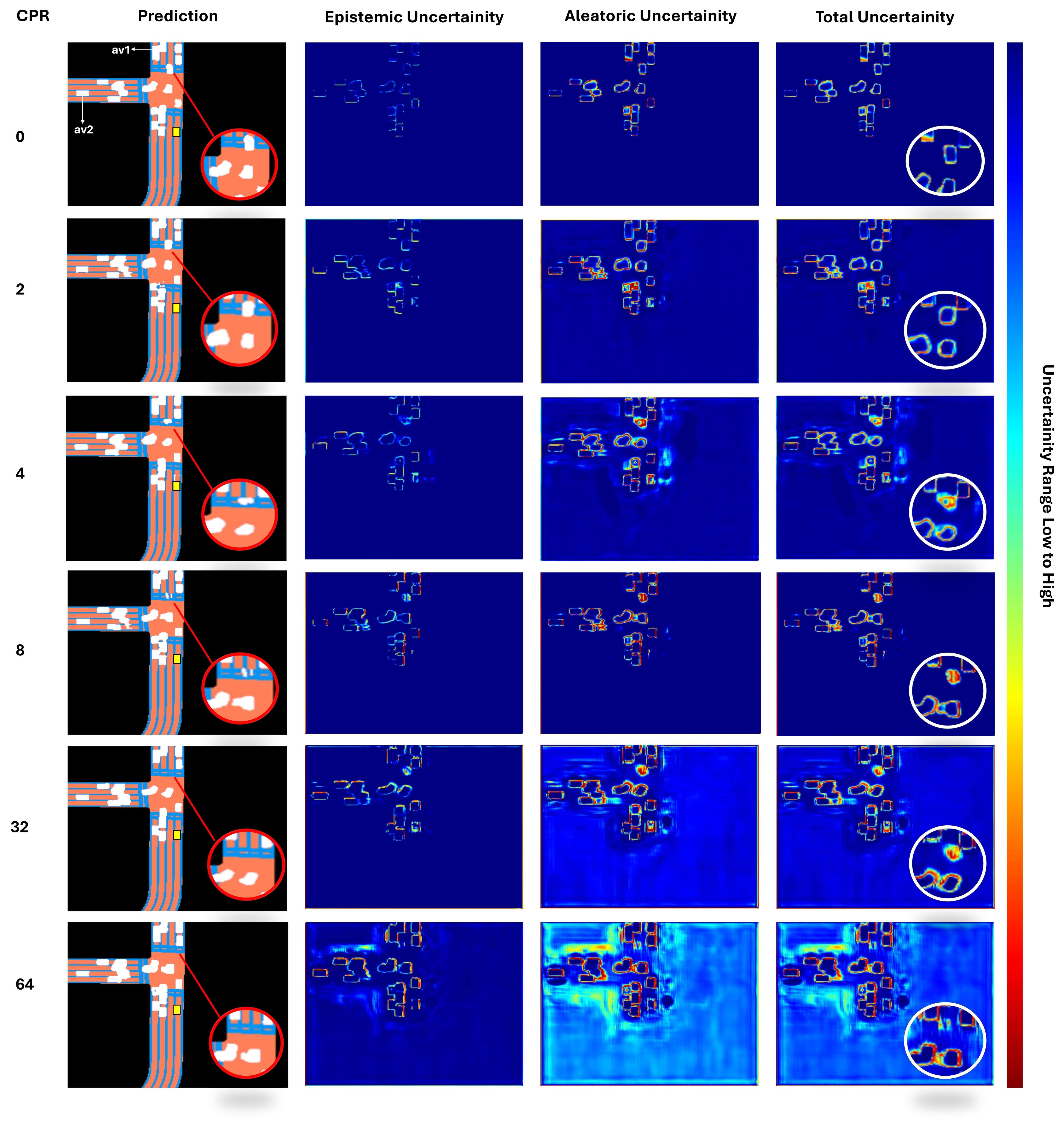}
    \caption{\small \textbf{Uncertainty estimation under varying compression rates.} As CPR increases (0→64), segmentation quality degrades (red circles). Our method produces uncertainty maps that effectively capture this degradation, with progressively higher epistemic and aleatoric uncertainty in vulnerable regions, demonstrating reliable uncertainty estimation under communication constraints.}
    \label{fig:visual_res2}
\end{figure*}

\subsection{Implementation Details}

Our experiments are built upon the CoBEVT~\cite{xu2022cobevt} architecture for collaborative perception. We first pre-train a single-vehicle variant of CoBEVT (referred to as SinBEVT) for 70~epochs. The corresponding pre-trained model is then used to initialize the primary cooperative perception network. For experiments involving uncertainty estimation with lower communication volume, we fine-tune the trained model with  different compression rates for 40~epochs. The training is performed with a batch size of~1 using an ELBO-style loss function \cite{krueger2017bayesian} that combines the standard segmentation loss, negative log-likelihood (NLL), and Kullback--Leibler (KL) divergence~\cite{Kullback1951_KLdivergence}. Specifically, the total loss is formulated as
\begin{equation}
\mathcal{L} = \mathcal{L}_{\text{seg}} + \lambda_{\text{NLL}} \, \mathcal{L}_{\text{NLL}} + \lambda_{\text{KL}} \, \mathcal{L}_{\text{KL}},
\end{equation}

where $\mathcal{L}_{\text{seg}}$ denotes the standard weighted cross-entropy segmentation loss, $\mathcal{L}_{\text{NLL}}$ encourages accurate probabilistic predictions, $\mathcal{L}_{\text{KL}}$ regularizes the posterior of the Bayesian hypernetwork, and $\lambda_{\text{NLL}}$ and $\lambda_{\text{KL}}$ are weighting coefficients. In our experiments, $K$ denotes the number of stochastic forward passes used for uncertainty estimation. For MC Dropout, we set $K=20$, while for HyperV2X, we use $K=10$. All experiments are carried out on an NVIDIA A100 with 80~GB memory.



\subsection{Results}
\subsubsection{Quantitative Results}

Table~\ref{tab:opv2v_segmentation_uncert} presents detailed quantitative results of various state-of-the-art methods on OPV2V camera track. Notably, our method is among the first to explore uncertainty estimation for cooperative BEV semantic segmentation while achieving comparatively state-of-the-art IoU. Furthermore, we evaluate the quality of our uncertainty estimation against baseline approaches, such as MC Dropout, in Tab.~\ref{tab:uncertainty_methods}. The results demonstrate that our proposed \mbox{Hyper-V2X} method not only achieves a higher segmentation accuracy (IoU) but also yields reliable uncertainty estimates, as evidenced by lower Expected Calibration Error (ECE), BS and NLL. 
Table~\ref{tab:compression_uncertainty} presents the effect of data compression rates on BEV semantic segmentation. As the communication volume (CV) is reduced, the IoU decreases, while ECE, BS and NLL exhibit a corresponding increase, indicating a degradation in predictive accuracy and calibration.

\subsubsection{Qualitative Results}
Fig.~\ref{fig:visual_res1} presents a visual comparison of our stochastic BEV segmentation predictions along with the corresponding uncertainty maps on the OPV2V benchmark. We visualize both epistemic and aleatoric uncertainty to provide comprehensive insight into the model's predictive confidence. Notably, high uncertainty is observed at the boundaries of semantic objects in both epistemic and aleatoric uncertainty maps. This pattern is particularly evident where geometric irregularities occur -- for instance, when predicted vehicle boundaries deviate from ground truth rectangular box. Regions with incomplete or occluded observations consistently exhibit elevated uncertainty, reflecting the model’s awareness of perceptual ambiguity under challenging visual conditions.

Fig.~\ref{fig:visual_res2} illustrates the impact of varying compression rates on prediction quality and uncertainty estimation for a representative scene. As the compression rate (CPR) increases from 0 to 64, we observe progressive degradation in segmentation performance. Specific objects that are accurately detected at compression rate 0 (highlighted in red circles) gradually deteriorate as communication bandwidth is reduced, until they are no longer detected at compression rate 64. Critically, our uncertainty maps effectively capture this degradation-regions (highlighted in white circles) and exhibits progressively higher epistemic and aleatoric uncertainty as compression increases. This demonstrates that \mbox{Hyper-V2X} provides reliable uncertainty estimates under communication constraints. Such uncertainty-aware predictions are particularly valuable in cooperative perception systems, where bandwidth limitations and communication losses are inherent challenges, enabling downstream tasks to appropriately weight or discard unreliable predictions.

\subsection{Ablation Study}
\label{subsec:ablation}

We evaluate the impact of V2X context embedding for conditioning Bayesian hypernetworks. Table~\ref{table:ablation_study_1} presents the performance of Hyper-V2X when conditioned on Gaussian noise. The results show decreased IoU and higher ECE, highlighting the benefit of V2X context embedding.
\begin{table}[h!]
\centering
\caption{\small Effect of Task embedding on Hyper-V2X }
\label{table:ablation_study_1}
\begin{tabular}{l S[table-format=2.1] S[table-format=1.4]}
  \toprule
    \textbf{Hyper-V2X} & \textbf{Vehicle IoU $\uparrow$} & \textbf{ECE $\downarrow$}\\
  \midrule
Gaussian Noise $\mathcal{N}(0, 0.01)$  & 58.4 & 0.0158  \\
With V2X context embedding  & 61.4 &  0.0140  \\
\bottomrule
\end{tabular}
\end{table}


\section{Conclusions}
\label{sec:concl}


This paper introduces Hyper-V2X, a hypernetwork-based framework for estimating epistemic and aleatoric uncertainties in cooperative BEV segmentation. By leveraging a partial weight generation scheme and conditioning  Bayesian hypernetwork on a V2X context embedding, Hyper-V2X enables efficient and scalable uncertainty estimation. Experiments on the OPV2V benchmark show that our method achieves accurate, well-calibrated uncertainty estimates and improves segmentation performance with minimal architectural overhead. Future work will extend Hyper-V2X to other cooperative perception tasks and explore uncertainty-guided communication and fusion strategies to enhance robustness and efficiency in connected autonomous systems.
\newline



\bibliography{IEEEabrv.bib, ref}

@STRING{IEEE_J_PAMI       = "{IEEE} Trans. Pattern Anal. Mach. Intell."}

@inproceedings{louizos2017multiplicative,
  title={Multiplicative normalizing flows for variational {Bayesian} neural networks},
  author={Louizos, Christos and Welling, Max},
  booktitleL={International Conference on Machine Learning},
  booktitle = 	 {Inter. Conf. on Mach. Learn. (ICML)},
  pagesL={2218--2227},
  year={2017},
  organization={PMLR},
  url={https://proceedings.mlr.press/v70/louizos17a/louizos17a.pdf}
}

@article{su2024collaborative,
  title={Collaborative multi-object tracking with conformal uncertainty propagation},
  authorL={Su, Sanbao and Han, Songyang and Li, Yiming and Zhang, Zhili and Feng, Chen and Ding, Caiwen and Miao, Fei},
  author={Su, Sanbao and others},
  journalLong={IEEE Robotics and Automation Letters},
  journal={{IEEE} Robot. Autom. L.},
  volume={9},
  number={4},
  pagesL={3323--3330},
  year={2024},
  publisherL={IEEE},
  note={doi:~10.1109/LRA.2024.3364450}
}

@inproceedings{chan2024estimating,
  title={Estimating epistemic and aleatoric uncertainty with a single model},
  author={Chan, Matthew and Molina, Maria and Metzler, Chris},
  booktitleL={Advances in Neural Information Processing Systems},
  booktitle = {Adv. Neural Inform. Process. Syst. (NeurIPS)},
  pagesL={109845--109870},
  year={2024},
  url={https://proceedings.neurips.cc/paper_files/paper/2024/file/c693c3ff83259aebcd55a41ab19a5d84-Paper-Conference.pdf}
}

@INPROCEEDINGS{Hu_CollaborHelps_CVPR_2023,
  authorL={Hu, Yue and Lu, Yifan and Xu, Runsheng and Xie, Weidi and Chen, Siheng and Wang, Yanfeng},
  author={Hu, Yue and others},
  booktitleL={IEEE/CVF Conference on Computer Vision and Pattern Recognition (CVPR)}, 
  booktitleSh={IEEE/CVF CVPR}, 
  booktitle={IEEE/CVF Conf. Comput. Vis. Pattern Recog. (CVPR)},
  title={Collaboration Helps Camera Overtake {LiDAR} in {3D} Detection}, 
  year={2023},
  pagesL={9243-9252},
  note={doi:~10.1109/CVPR52729.2023.00892}}

@inproceedings{xu2022v2x,
  title={{V2X-ViT}: Vehicle-to-everything cooperative perception with vision transformer},
  authorLong={Xu, Runsheng and Xiang, Hao and Tu, Zhengzhong and Xia, Xin and Yang, Ming-Hsuan and Ma, Jiaqi},
  author={Xu, Runsheng and others},
  booktitleL={European Conference on Computer Vision (ECCV)},
  booktitle={Eur. Conf. Comput. Vis. (ECCV)},
  pagesL={107--124},
  year={2022},
  organization={Springer},
  note={doi:~10.1007/978-3-031-19842-7\_7}
}

@INPROCEEDINGS{xu2022cobevt,
  title={{CoBEVT}: Cooperative bird's eye view semantic segmentation with sparse transformers},
  authorL={Xu, Runsheng and Tu, Zhengzhong and Xiang, Hao and Shao, Wei and Zhou, Bolei and Ma, Jiaqi},
  author={Xu, Runsheng and others},
  booktitle={Conference on Robot Learning (CoRL)},
  journalL={arXiv preprint arXiv:2207.02202},
  url={https://proceedings.mlr.press/v205/xu23a/xu23a.pdf},
  year={2022}
}

@INPROCEEDINGS{Fu_GenerMaps_CVPR_2025,
  authorL={Fu, Jiahui and Gong, Yue and Wang, Luting and Zhang, Shifeng and Zhou, Xu and Liu, Si},
  author={Fu, Jiahui and others},
  booktitleL={IEEE/CVF Conference on Computer Vision and Pattern Recognition (CVPR)}, 
  booktitle={IEEE/CVF Conf. Comput. Vis. Pattern Recog. (CVPR)},
  booktitleSh={IEEE/CVF CVPR}, 
  title={Generative Map Priors for Collaborative {BEV} Semantic Segmentation}, 
  year={2025},
  pagesL={11919-11928},
  note={doi:~10.1109/CVPR52734.2025.01113}}

@INPROCEEDINGS{Jagtap_V2XGaussians_2025,
  author={Jagtap, Abhishek Dinkar and Song, Rui and Sadashivaiah, Sanath Tiptur and Festag, Andreas},
  booktitleL={2025 IEEE Intelligent Vehicles Symposium (IV)}, 
  booktitle={2025 IEEE IV Symposium}, 
  title={{V2X-Gaussians}: Gaussian Splatting for Multi-Agent Cooperative Dynamic Scene Reconstruction}, 
  year={2025},
  pagesL={1033-1039},
  note={doi:~10.1109/IV64158.2025.11097436}}

@inproceedings{ha2017hypernetworks,
  title={HyperNetworks},
  author={David Ha and Andrew M. Dai and Quoc V. Le},
  booktitleSHort={ICLR},
  booktitleL={International Conference on Learning Representations (ICLR)},
  booktitle={Int. Conf. Learn. Represent. (ICLR)},
  year={2017},
  urlALT={https://openreview.net/forum?id=rkpACe1lx},
  note={doi:~10.48550/arXiv.1609.09106},
}

@inproceedings{krueger2017bayesian,
  title={Bayesian hypernetworks},
  authorLong={Krueger, David and Huang, Chin-Wei and Islam, Riashat and Turner, Ryan and Lacoste, Alexandre and Courville, Aaron},
  author={Krueger, David and others},
  year={2017},
  noteL={arXiv preprint arXiv:1710.04759},
  booktitle={NIPS Workshop Bayesian Deep Learning},
  url = {https://bayesiandeeplearning.org/2017/papers/34.pdf},

}

@inproceedings{hemati2023partial,
  title={Partial hypernetworks for continual learning},
  author={Hemati, Hamed and Lomonaco, Vincenzo and Bacciu, Davide and Borth, Damian},
  booktitle={Conference on Lifelong Learning Agents (CoLLAs)},
  pagesLong={318--336},
  year={2023},
  organizationLong={PMLR},
  url = {https://proceedings.mlr.press/v232/hemati23a.html},
}

@ARTICLE{liu2024:collaborative,
  authorLong={Liu, Genjia and Hu, Yue and Xu, Chenxin and Mao, Weibo and Ge, Junhao and Huang, Zhengxiang and Lu, Yifan and Xu, Yinda and Xia, Junkai and Wang, Yafei and Chen, Siheng},
  author={Genjia Liu and others},
  journalL={IEEE Transactions on Pattern Analysis and Machine Intelligence}, 
  journal=IEEE_J_PAMI,
  title={Toward Collaborative Autonomous Driving: Simulation Platform and End-to-End System}, 
  year={2025},
  volume={47},
  number={8},
  pagesL={6566-6584},
  note={doi:~10.1109/TPAMI.2025.3560327}}

@INPROCEEDINGS{Su2022uncertainty,
  author = {Su, Sanbao and others},
  authorLong = {Su, Sanbao and Li, Yiming and He, Sihong and Han, Songyang and Feng, Chen and Ding, Caiwen and Miao, Fei},
  title = {Uncertainty Quantification of Collaborative Detection for Self-Driving},
  year = {2023},
  booktitleL = {IEEE International Conference on Robotics and Automation (ICRA)},
  booktitle = {IEEE Inter. Conf. on Rob. and Autom (ICRA)},
  booktitleSh = {ICRA},
  organizationL = {IEEE},
  note = {doi:~10.1109/ICRA48891.2023.10160367}
}

@inproceedings{Hu2022_Where2comm,
  title = {Where2comm: Communication-Efficient Collaborative Perception via Spatial Confidence Maps},
  shorttitle = {Where2comm},
  author = {Hu, Yue and Fang, Shaoheng and Lei, Zixing and Zhong, Yiqi and Chen, Siheng},
  year = {2022},
  monthL = sep,
  booktitleL = {Neural Information Processing Systems (NeurIPS)},
  booktitle = {Adv. Neural Inform. Process. Syst. (NeurIPS)},
  booktitleSh = {NeurIPS},
  url = {https://proceedings.neurips.cc/paper\_files/paper/2022/file/1f5c5cd01b864d53cc5fa0a3472e152e-Paper-Conference.pdf},
  urltiny={https://tinyurl.com/mr33y3fc},
  noteLLL={doi:~10.48550/arXiv.2209.12836}
}

@inproceedings{Yang2023_How2comm,
  title = {How2comm: {Communication}-Efficient and Collaboration-Pragmatic Multi-Agent Perception},
  authorL = {Yang, Dingkang and Yang, Kun and Wang, Yuzheng and Liu, Jing and Xu, Zhi and Yin, Rongbin and Zhai, Peng and Zhang, Lihua},
  author = {Yang, Dingkang and others},
  booktitleL = {Neural Information Processing Systems (NeurIPS)},
  booktitle = {Adv. Neural Inform. Process. Syst. (NeurIPS)},
  booktitleSh = {NeurIPS},
  year = {2023},
  addressL = {New Orleans, LA, USA},
  url = {https://proceedings.neurips.cc/paper\_files/paper/2023/file/4f31327e046913c7238d5b671f5d820e-Paper-Conference.pdf},
  urltiny={https://tinyurl.com/5ymzcbth}
}

@INPROCEEDINGS{Chen2019_Cooper,
  author={Chen, Qi and Tang, Sihai and Yang, Qing and Fu, Song},
  booktitle={IEEE Intern. Conf. on Distr. Comp. Syst. (ICDCS)}, 
  booktitleL={IEEE ICDCS}, 
  title={Cooper: Cooperative Perception for Connected Autonomous Vehicles Based on {3D} Point Clouds}, 
  year={2019},
  pagesL={514-524},
  note={doi:~10.1109/ICDCS.2019.00058}
}

@inproceedings{song2024_collaborative,
  title={Collaborative Semantic Occupancy Prediction with Hybrid Feature Fusion in Connected Automated Vehicles},
  authorL={Song, Rui and Liang, Chenwei and Cao, Hu and Yan, Zhiran and Zimmer, Walter and Gross, Markus and Festag, Andreas and Knoll, Alois},
  author={Song, Rui and others},
  booktitleLong={IEEE/CVF International Conference on Computer Vision and Pattern Recognition (CVPR)},
  booktitleSh={IEEE/CVF CVPR},
  booktitle={IEEE/CVF Conf. Comput. Vis. Pattern Recog. (CVPR)},
  year={2024},
  note={doi:~10.1109/CVPR52733.2024.01704}
 }

@INPROCEEDINGS{xu2022opv2v,
  authorLong={Xu, Runsheng and Xiang, Hao and Xia, Xin and Han, Xu and Li, Jinlong and Ma, Jiaqi},
  author={Runsheng Xu and others},
  booktitleL={IEEE International Conference on Robotics and Automation (ICRA)}, 
  booktitle = {IEEE Inter. Conf. on Rob. and Autom (ICRA)},
  organizationL = {IEEE},
  title={{OPV2V}: An Open Benchmark Dataset and Fusion Pipeline for Perception with Vehicle-to-Vehicle Communication}, 
  year={2022},
  pagesL={2583-2589},
  note={doi:~10.1109/ICRA46639.2022.9812038}
}

@INPROCEEDINGS{Delooz2023_CollPerception,
  author={Delooz, Quentin and Festag, Andreas and Vinel, Alexey and Lobo, Silas C.},
  booktitleL={2023 IEEE Intelligent Vehicles Symposium (IV)},
  booktitle={IEEE IV Symposium},
  title={Simulation-based Performance Optimization of {V2X} Collective Perception by Adaptive Object Filtering}, 
  year={2023},
  pagesL={1-8},
  note={doi:~10.1109/IV55152.2023.10186788}}

@INPROCEEDINGS{FirstMile,
  authorL={Song, Rui and Festag, Andreas and Jagtap, Abhishek Dinkar and Bialdyga, Maximilian and Yan, Zhiran and Otte, Maximilian and Sadashivaiah, Sanath Tiptur and Knoll, Alois},
  author={Song, Rui and others},
  booktitle={IEEE IV Symposium}, 
  title={{First} {Mile}: An Open Innovation Lab for Infrastructure-Assisted Cooperative Intelligent Transportation Systems}, 
  year={2024},
  pagesL={1635-1642},
  note={doi:~10.1109/IV55156.2024.10588500}}

@inproceedings{NIPS2017_9ef2ed4b,
 author = {Lakshminarayanan, Balaji and Pritzel, Alexander and Blundell, Charles},
 booktitleL = {Advances in Neural Information Processing Systems (NeurIPS)},
 booktitle = {Adv. Neural Inform. Process. Syst. (NeurIPS)},
 booktitleSh = {NIPS},
 editorLong = {I. Guyon and U. Von Luxburg and S. Bengio and H. Wallach and R. Fergus and S. Vishwanathan and R. Garnett},
 publisherLong = {Curran Associates, Inc.},
 title = {Simple and Scalable Predictive Uncertainty Estimation using Deep Ensembles},
 url = {https://proceedings.neurips.cc/paper_files/paper/2017/file/9ef2ed4b7fd2c810847ffa5fa85bce38-Paper.pdf},
 volumeL = {30},
 year = {2017}
}

@InProceedings{pmlr-v48-gal16,
  title = 	 {Dropout as a Bayesian Approximation: Representing Model Uncertainty in Deep Learning},
  author = 	 {Gal, Yarin and Ghahramani, Zoubin},
  booktitleL = 	 {ICML},
  booktitleLL = 	 {International Conference on Machine Learning (ICML)},
  booktitle = 	 {Inter. Conf. on Mach. Learn. (ICML)},
  year = 	 {2016},
  volumeL = 	 {48},
  monthL = 	 {20--22 Jun},
  url = 	 {https://proceedings.mlr.press/v48/gal16.html},
}

@inproceedings{chen2019fcooper, 
  authorL = {Chen, Qi and Ma, Xu and Tang, Sihai and Guo, Jingda and Yang, Qing and Fu, Song},
  author = {Chen, Qi and others},
  title = {{F-cooper}: Feature based cooperative perception for autonomous vehicle edge computing system using {3D} point clouds}, 
  year = {2019}, 
  note = {doi:~10.1145/3318216.3363300}, 
  booktitle = {ACM/IEEE Symposium on Edge Computing}, 
  pagesL = {88–100}, 
  numpagesL = {13}, 
}

@inproceedings{wang2020v2vnet,
  title={{V2VNet:}~{Vehicle}-to-vehicle communication for joint perception and prediction},
  authorLONG={Wang, Tsun-Hsuan and Manivasagam, Sivabalan and Liang, Ming and Yang, Bin and Zeng, Wenyuan and Urtasun, Raquel},
  author={Wang, Tsun-Hsuan and others},
  booktitleL={European Conference on Computer Vision (ECCV)},
  booktitle={Eur. Conf. Comput. Vis. (ECCV)},
  pagesL={605--621},
  addressL={Glasgow, UK},
  month=aug,
  year={2020},
  organization={Springer},
  note={{DOI:}~10.1007/978-3-030-58536-5\_36}
}

@inproceedings{li2021learning,
  title={Learning distilled collaboration graph for multi-agent perception},
  author={Li, Yiming and others},
  authorLONG={Li, Yiming and Ren, Shunli and Wu, Pengxiang and Chen, Siheng and Feng, Chen and Zhang, Wenjun},
  booktitleL={Advances in Neural Information Processing Systems (NeurIPS)},
  booktitle = {Adv. Neural Inform. Process. Syst. (NeurIPS)},
  volumeL={34},
  pagesL={29541--29552},
  year={2021},
  url={https://proceedings.neurips.cc/paper_files/paper/2021/hash/f702defbc67edb455949f46babab0c18-Abstract.html}
}

@ARTICLE{Wan_SystematicLitRev_TITS_2025,
  authorLong={Lei Wan and Jianxin Zhao and Andreas Wiedholz and Manuel Bied and Mateus Martinez de Lucena and Abhishek Dinkar Jagtap and Andreas Festag and Antônio Augusto Fröhlich and Hannan Ejaz Keen and Alexey Vinel},
  author={Lei Wan and others},
  journalL={Accepted for IEEE Transactions on Intelligent Transportation Systems}, 
  journal={Accepted for {IEEE} Trans. Intell. Transp. Syst.},
  title={Systematic Literature Review on Vehicular Collaborative Perception -- {A} Computer Vision Perspective}, 
  year={2025},
  note={doi:~10.48550/arXiv.2504.04631},
}

@article{Huang_UncertCollPerc_Adv_2025,
  author={Huiqun Huang and Cong Chen and Jean-Philippe Monteuuis and Jonathan Petit and Fei Miao},
  title={Uncertainty Quantification for Collaborative Object Detection Under Adversarial Attacks},
  year={2025},
  monthL={February},
  journal={arXiv:2502.02537}
}

@inproceedings{Yang2023UncertaintyBEV,
  author    = {Bing Yang and others},
  title     = {Evaluating Uncertainty Quantification for Bird’s Eye View Semantic Segmentation},
  booktitle = {Workshop on Uncertainty Reasoning and Quantification in Decision Making},
  year      = {2023},
  url       = {https://charliezhaoyinpeng.github.io/UDM-KDD23/ap/}
}

@ARTICLE{Chang2023_BEV-V2X,
  authorL={Chang, Cheng and Zhang, Jiawei and Zhang, Kunpeng and Zhong, Wenqin and Peng, Xinyu and Li, Shen and Li, Li},
  author={Chang, Cheng and others},
  journal={IEEE Transactions on Intelligent Vehicles}, 
  title={{BEV-V2X}: Cooperative Birds-Eye-View Fusion and Grid Occupancy Prediction via V2X-Based Data Sharing}, 
  year={2023},
  volume={8},
  number={11},
  pagesL={4498-4514},
  note={doi:10.1109/TIV.2023.3293954}
}

@article{Kullback1951_KLdivergence,
  author    = {Solomon Kullback and Richard A. Leibler},
  title     = {On Information and Sufficiency},
  journal   = {The Annals of Mathematical Statistics},
  volume    = {22},
  number    = {1},
  pages     = {79--86},
  year      = {1951},
  note       = {doi:~10.1214/aoms/1177729694}
}

@inproceedings{Blundell2015VI,
  author    = {Charles Blundell and Julien Cornebise and Koray Kavukcuoglu and Daan Wierstra},
  title     = {Weight Uncertainty in Neural Networks},
  booktitleL = {Proceedings of the 32nd International Conference on Machine Learning (ICML)},
  booktitle = 	 {Inter. Conf. on Mach. Learn. (ICML)},
  pagesL     = {1613--1622},
  year      = {2015},
  url       = {http://proceedings.mlr.press/v37/blundell15.html}
}

@ARTICLE{li2022bevformer,
  authorL={Li, Zhiqi and Wang, Wenhai and Li, Hongyang and Xie, Enze and Sima, Chonghao and Lu, Tong and Yu, Qiao and Dai, Jifeng},
  author={Li, Zhiqi and others},
  journalL={IEEE Transactions on Pattern Analysis \& Machine Intelligence},
  journal=IEEE_J_PAMI, 
  title={{BEVFormer}: Learning Bird’s-Eye-View Representation From {LiDAR}-Camera via Spatiotemporal Transformers},
  year={2025},
  volume={47},
  number={03},
  pagesL={2020-2036},
  note={doi:~10.1109/TPAMI.2024.3515454},
  monthLong=mar}

@InProceedings{Pavlitska_2025_ICCV,
    author    = {Pavlitska, Svetlana and Keskin, Beyza and Fa{\ss}bender, Alwin and Hubschneider, Christian and Z\"ollner, J. Marius},
    title     = {Extracting Uncertainty Estimates from Mixtures of Experts for Semantic Segmentation},
    booktitle = {Proceedings of the IEEE/CVF International Conference on Computer Vision (ICCV) Workshops},
    month     = {October},
    year      = {2025},
    pages     = {311-320}
}
\bibliographystyle{IEEEtran}


\end{document}